\documentclass[twoside]{romjist}
\usepackage[top=4.3cm, bottom=5.7cm, left = 3.74cm, right = 3.75cm]{geometry}
\usepackage{times}
\usepackage{authblk}
\usepackage{graphicx} 
\usepackage{amsmath}
\usepackage{amssymb}
\usepackage{amsthm}
\usepackage{cite}
\usepackage{graphicx}
\usepackage{epstopdf}
\usepackage{amssymb}
\usepackage{booktabs}
\usepackage{chemfig}
\usepackage{centernot} 
\usepackage[labelsep=space]{caption}

\begin{document}
      
    \issuevolume{X}
    \issuenumber{X}
    \issueyear{XXXX}
    \issuepages{XXX--XXX}

    \headauthors{S. Yin, L. Jiang}
    \headtitle{Confidence-weighted Fusion for Zero-shot Image Classification}
    
    \title{Multi-method Integration with Confidence-based Weighting for Zero-shot Image Classification}

    \author[1, *]{Siqi \uppercase{Yin}}
    \author[2]{Lifan \uppercase{Jiang}}

    \affil[1]{{\small School of Computer Science and Technology, Shandong University of Science and Technology, 579 Qianwan Gang Road, Huangdao District, Qingdao, 266590, Shandong, China}}
    \affil[2]{{\small School of Computer Science and Technology, Shandong University of Science and Technology, 579 Qianwan Gang Road, Huangdao District, Qingdao, 266590, Shandong, China}}
    \affil[ ]{ {\small \texttt {Email: siqiyin123@gmail.com$^*$,siqiyin@sdust.edu.cn$^*$, lifanjiang@sdust.edu.cn}}}
    \affil[ ]{{ \small $^*$ Corresponding author}}
    

    \maketitle
    
    \begin{abstract} This paper introduces a novel framework for zero-shot learning (ZSL), i.e., to recognize new categories that are unseen during training, by using a multi-model and multi-alignment integration method. Specifically, we propose three strategies to enhance the model's performance to handle ZSL: 1) Utilizing the extensive knowledge of ChatGPT and the powerful image generation capabilities of DALL-E to create reference images that can precisely describe unseen categories and classification boundaries, thereby alleviating the information bottleneck issue; 2) Integrating the results of text-image alignment and image-image alignment from CLIP, along with the image-image alignment results from DINO, to achieve more accurate predictions; 3) Introducing an adaptive weighting mechanism based on confidence levels to aggregate the outcomes from different prediction methods. Experimental results on multiple datasets, including CIFAR-10, CIFAR-100, and TinyImageNet, demonstrate that our model can significantly improve classification accuracy compared to single-model approaches, achieving AUROC scores above 96\% across all test datasets, and notably surpassing 99\% on the CIFAR-10 dataset.\end{abstract}


\section{Introduction}

\hspace{0.5cm} In recent years, deep learning technologies have made significant progresses, particularly in the field of image classification, where they have driven substantial advancements and achieved remarkable improvements in effectiveness. For instance, ResNet \cite{55} introduced a residual learning mechanism to address the difficulties in training deep networks, substantially increasing classification accuracy.Meanwhile, MobileNet \cite{57} achieved efficient computation on mobile and embedded devices with its lightweight deep separable convolution design. However, despite the significant success, these methods still suffer from some limitations, such as the typical requirement for the test categories to be the same as the training categories, which imposes restrictions in their practical applications. In real-world scenarios, we often encounter a large number of new categories that may not be included in the original training set. However, we may hope that these new classes can still be handled effectively by the model that has not been trained on them. In these cases, the aforementioned methods such as the vanilla ResNet, which are only able to classify the categories already been trained, would be limited in practicality and are thus difficult to be effectively deployed in real-world contexts.

To overcome the limitations of traditional image classification methods when dealing with unseen categories, Zero-Shot Learning (ZSL) has garnered increasing attention in recent years. ZSL aims to enable models to recognize categories that they have never seen during the training phase, thus being able to identifying new classes by leveraging the semantic relationships between categories. Several prominent ZSL models have been proposed and introduced progresses for this filed from various aspects. For example, an early ZSL model \cite{54} improved the classification accuracy of unseen bird species by learning the mapping between visual features and textual descriptions. Due to the relatively small number of images for training models, these methods still face challenges in achieving very satisfactory zero-shot generalization capabilities. How to further enhance the novel-class generalization ability of the zero-shot models is still a critical problem that needs to be addressed.

To address these challenges, CLIP (Contrastive Language-Image Pre-Training), an innovative image classification model, has emerged, excelling in unsupervised or zero-shot image classification tasks by leveraging the synergy between vision and language. Through extensive contrastive learning with large-scale image and text data, CLIP learns a universal visual-linguistic model, enabling direct zero-shot classification for new categories by aligning image features with textual descriptions. Specifically, CLIP evaluates the similarity between an image and a series of candidate text descriptions, allowing it to classify unseen categories without the need for additional training data. However, using text-image alignment for Zero-Shot Classification still poses challenges, such as semantic gap issues and contextual ambiguity problems, which limit the model's performance on complex or fine-grained categories. Therefore, there is an urgent need to develop new methods that can improve existing technologies to address these challenges, thus further enhancing the effectiveness and generalizability of Zero-Shot Learning in practical applications.

In this paper, we propose a novel framework designed to leverage the alignment of images with text as well as the alignment among images themselves, integrating multiple models such as CLIP and DINO to achieve a more robust and effective processing ability. In this framework, we propose several novel strategies including: 1) Utilizing the vast knowledge contained in the ChatGPT and the strong image generation capabilities of DALL-E to create reference images that can accurately describe classification boundaries, thereby aiding zero-shot classification where we have no access to the images for the novel class and thus easily suffer from the problem of information bottleneck; 2) We merge the text-image alignment and image-image alignment results of CLIP with the image-image alignment result of DINO to achieve more accurate predictions; 3) We also introduce an adaptive weighting mechanism based on confidence to integrate the results from different predictive models, which we validated across multiple datasets.

With the carefully designed integration method, our proposed model achieves remarkable effectiveness in handling zero-shot classification problems. To evaluate the effectiveness of our method, we conduct experiments on multiple datasets including CIFAR-10, CIFAR-100, and TinyImageNet. In comparison to conventional single models like the text-image alignment, image-image alignment of CLIP, and the image-image alignment of DINO, our integrated model has realized significant gains in classification accuracy AC, with improvements of 11.97\%, 26.06\%, and 12.6\% on CIFAR-10; 24.48\%, 42.48\%, and 5.42\% on CIFAR-100; and 32.32\%, 41.96\%, and 1.2\% on TinyImageNet, respectively. Furthermore, our model score above 96\% on the AUROC indicator across all test datasets, particularly exceeding 99\% on the CIFAR-10 dataset. These excellent outcomes not only highlight the advantages of our method in classification accuracy but also prove its exceptional ability to distinguish both between open and closed set categories. These experimental results robustly support the multi-model fusion strategy we propose. By integrating the strengths of different models, our approach has not only improved the recognition accuracy of known categories but, more importantly, it has exhibited a significant improvement in the generalization ability to unseen categories. This confirms that a multi-model fusion strategy is an effective path when constructing models capable of handling broad and complex visual tasks.




\section{Related Work}
\subsection{Zero-Shot Classification}
\hspace{0.5cm} Zero-shot classification (ZSC) is a machine learning paradigm aimed at enabling models to recognize categories they have not seen during the training phase. Unlike traditional supervised learning methods, ZSC does not rely on direct experience with every class but rather achieves classification through understanding shared knowledge or attributes among categories. This capability is crucial for dealing with data scarcity or category diversity issues, especially in fields such as Natural Language Processing (NLP) and Computer Vision (CV). In recent years, significant progress is made in the field of ZSC with the advancement of deep learning technologies. In particular, pre-trained language models such as GPT-3 \cite{1} and BERT \cite{2} show great potential in ZSC tasks. These models, having been pre-trained on massive text datasets, learn rich linguistic features and world knowledge, enabling them to infer unknown categories without explicit examples. Additionally, Graph Neural Networks (GNNs) prove to be effective tools for addressing ZSC problems. By operating on graph-structured data, GNNs capture complex relationships and attributes between categories, offering new avenues for accurate classification of unseen classes \cite{3,4}. Despite progress, ZSC still faces many challenges \cite{8,9}. One such challenge is how to enhance the model's generalization to new domains and new categories, requiring the model to not only capture commonalities between categories but also adapt to those situations significantly different from the training data distribution. Deep Calibration Networks (DCN) proposed in \cite{7} aim to reduce the uncertainty of models when dealing with unseen categories. Research in \cite{8} explores how meta-learning can help models perform better in domains they have not seen. A framework based on Generative Adversarial Networks (GANs) is proposed in \cite{9}, which can generate features representative of unseen categories, thus improving performance on ZSC tasks. Moreover, data imbalance and attribute sparsity are key issues that need to be addressed \cite{11}. The Balanced Meta-Softmax method proposed in \cite{11} is also of referential value for handling data imbalance issues in ZSC. This paper proposes a multi-model fusion approach that uses the capabilities of the DALL-E large model to generate images of unseen categories and visually similar pictures. Additionally, it applies softmax method processing in feature extraction to alleviate the ZSC classification problem.
\subsection{Clip}
CLIP (Contrastive Language–Image Pre-training) is a model developed by OpenAI that is pre-trained on a vast array of images and text data to improve generalization in computer vision tasks. The CLIP model is first introduced in \cite{13}, demonstrating its performance across multiple computer vision benchmarks, particularly showing its powerful capability in zero-shot classification tasks. It highlights the potential of learning visual models with natural language supervision, that is, training using widely available image-text pairs on the internet, enabling the model to connect visual content with descriptive text. Due to its strong generalization abilities, CLIP is applied in various domains, including but not limited to: (1) Zero-Shot Classification \cite{14, 25, 26}: CLIP can classify specific categories of images without having seen them before, exhibiting remarkable adaptability when dealing with novel objects or concepts. Research in \cite{14} utilizes the CLIP model to improve the performance of Model-Agnostic Zero-Shot Classification (MA-ZSC) by increasing the diversity of synthetic training images. Modifications in the text-to-image generation process enhance image diversity and show significant performance improvements across various classification architectures, rivaling state-of-the-art models like CLIP. (2) Image Retrieval \cite{15, 29, 31}: By retrieving relevant images through text descriptions, CLIP can understand complex queries and return matching pictures, which is particularly useful in search engines and digital image library management. \cite{15} proposes an FND-CLIP, a CLIP-based multimodal fake news detection framework that extracts deep representations from images and text and utilizes features generated by CLIP for fusion to enhance the capability of fake news detection. \cite{31} presents a model that expands the CLIP model, CLIP-ITA, by adding extra encoders and projection heads to adapt to category-to-image retrieval tasks in e-commerce. (3) Multimodal Learning \cite{16, 32, 15}: The design philosophy of CLIP fosters cross-learning between vision and language, which is crucial for enhancing machines' understanding and generation capabilities for multimodal content, such as videos and text-image content. \cite{16} modifies CLIP's encoders for prompt learning with video data, achieving a balance between capturing spatio-temporal cues and avoiding extensive fine-tuning. This paper capitalizes on the powerful visual feature extraction of CLIP, its capability for image-text matching, and its exceptional generalization ability. Building on the foundation of CLIP, we perform image-to-image matching and leverage the alignment verification capability of CLIP to assist DALL-E in generating images, aiming to further explore and research the Zero-Shot Classification problem.

\subsection{Dino}
DINO (Distillation with NO labels) is a self-supervised learning algorithm primarily utilized for visual representation learning. This approach employs knowledge distillation techniques to train visual Transformer networks without the need for labels. DINO leverages its own outputs as a supervisory signal by comparing different versions of images, such as varying sizes or those processed through data augmentation, to learn useful visual representations. The key to this method lies in generating a feature representation that is sensitive to the image content while being robust to changes in viewpoint, size, and other variations.
The DINO algorithm is first detailed in \cite{17}, where it is shown to excel in multiple visual representation learning tasks, particularly in the domain of self-supervised learning. By training visual Transformers in a self-supervised manner, it demonstrates performance on par with or even superior to supervised learning methods. Due to its powerful visual representation learning capabilities, DINO is applied in multiple domains, including: (1) Image Classification \cite{17, 19}: Although DINO is a self-supervised learning method, the representations learned through it can be utilized for downstream tasks such as image classification, typically completed by adding a simple classifier on top of the DINO-learned representations. Research in \cite{17} employs a k-NN classifier and discovers that self-supervised ViT features contain explicit information about semantic segmentation of images, which is less apparent in supervised ViT or convolutional networks. (2) Object Detection and Segmentation \cite{19, 36}: The feature representations learned by DINO are also very sensitive to the location and boundaries of objects within images, making it suitable for object detection and segmentation tasks. Work in \cite{19} introduces Mask DINO, a unified Transformer-based object detection and segmentation framework that significantly improves performance across all segmentation tasks and sets new records for instance, panoramic, and semantic segmentation. In this paper, we leverage DINO's ability to capture rich features of images without the need for labels, conducting checks on the alignment between images and images.

\subsection{GPT}
GPT (Generative Pre-trained Transformer) is a large-scale natural language processing model developed by OpenAI. It is first introduced in \cite{20}, detailing a method that combines unsupervised pre-training with supervised fine-tuning to enhance language understanding capabilities. Currently, GPT evolves to its fourth iteration, GPT-4\cite{39}, which is not only capable of text generation but can also perform complex tasks such as question answering, summarization, and translation \cite{39, 40}. GPT-4 is able to understand and generate text that is near human-level, marking a significant breakthrough in the field of language models. With extensive data pre-training, it learns a wide range of language patterns and knowledge, enabling it to excel in various tasks such as writing assistance, conversational systems, and content creation.
In addition to natural language processing (NLP) tasks, GPT's applications extend to but are not limited to several fields, including healthcare analytics \cite{42, 43}, education \cite{46, 47}, legal consultancy \cite{48}, and artistic creation \cite{21, 50}. Particularly in the realm of artistic creation, DALL-E \cite{21}, built upon similar generative pre-training technology, can generate images based on textual descriptions provided by users. This process demonstrates the advanced capabilities of artificial intelligence in image recognition and generation. In this paper, we utilize the powerful question-answering and text generation capabilities of GPT along with the image-generating ability of DALL-E to create images for negative classes (unseen categories). Moreover, we generate images for all potential categories based on each category's unique visual features, serving as a predictive reference for subsequent classification tasks.

\subsection{Data Augmentation}
Data augmentation techniques play a crucial role in the field of machine learning, augmenting training datasets and introducing diversity to enhance model generalization and performance. Traditional data augmentation methods, such as random cropping, rotation, scaling, and flipping, are widely adopted in computer vision tasks. In recent years, advanced data augmentation strategies have been proposed to further strengthen model robustness and generalization. Techniques like CutMix \cite{64} aim to generate diversified training samples by blending multiple images, thereby enhancing performance in tasks such as object detection and semantic segmentation. Data augmentation techniques have also been tailored to specific domains and tasks, including natural language processing (NLP), medical imaging, and audio processing. For example, common techniques in NLP tasks include backtranslation and paraphrasing, used to augment text data \cite{67}.In conclusion, data augmentation techniques are integral components of modern machine learning workflows, evolving from traditional augmentation methods to advanced strategies and domain-specific approaches. Researchers have demonstrated the effectiveness of data augmentation in improving model performance, robustness, and generalization across various domains and applications. Our data augmentation method introduces innovative improvements by leveraging the extensive knowledge of ChatGPT and the powerful image generation capability of DALL-E to create precise reference images that describe unseen categories and classification boundaries. This allows us to acquire reference samples for new categories and those situated at classification boundaries. We also enable the generated images to have similar appearances as their confusing categories, thus helping to learn a clearer classification boundary between different classes. This makes it significantly different from the above-mentioned traditional methods \cite{64,67}. This methodology significantly enhances model performance, particularly in handling unseen or ambiguous categories. \cite{82} also employs generative models to generate images for helping classify unseen classes. However, the mechanism of our method is significantly different. We utilize the extensive knowledge of ChatGPT and the powerful image generation capabilities of DALLE to create reference images that precisely describe unseen categories and classification boundaries, obtaining reference samples for new categories and those at the classification boundaries. This differs from the approach by \cite{82}, which aimed to augment training samples using DALLE. Unlike the preprocessing done during training in previous methods, our approach supplements high-quality boundary images during prediction as reference images. Additionally, we propose leveraging CLIP’s image feature extraction capability, integrating not only its text-image alignment but also image-image alignment techniques during classification. Compared to models without this technique, ours achieved a respective increase of 0.45\%, 0.24\%, and 0.23\% in classification accuracy on the CIFAR10, CIFAR100, and TinyImageNet datasets. Furthermore, we introduce a novel confidence-based adaptive weighting mechanism to aggregate results from different prediction models that is not utilized in \cite{82}. This mechanism allows us to weigh the results of each prediction model based on its confidence level, enabling more effective integration of multiple model predictions. Such adaptive weighting enhances the flexibility of our approach to accommodate varying performances of different models in diverse scenarios, further improving overall prediction accuracy and robustness.
\begin{figure*}[t]
    \centering
    \includegraphics[width=1.0\linewidth,height=210pt]{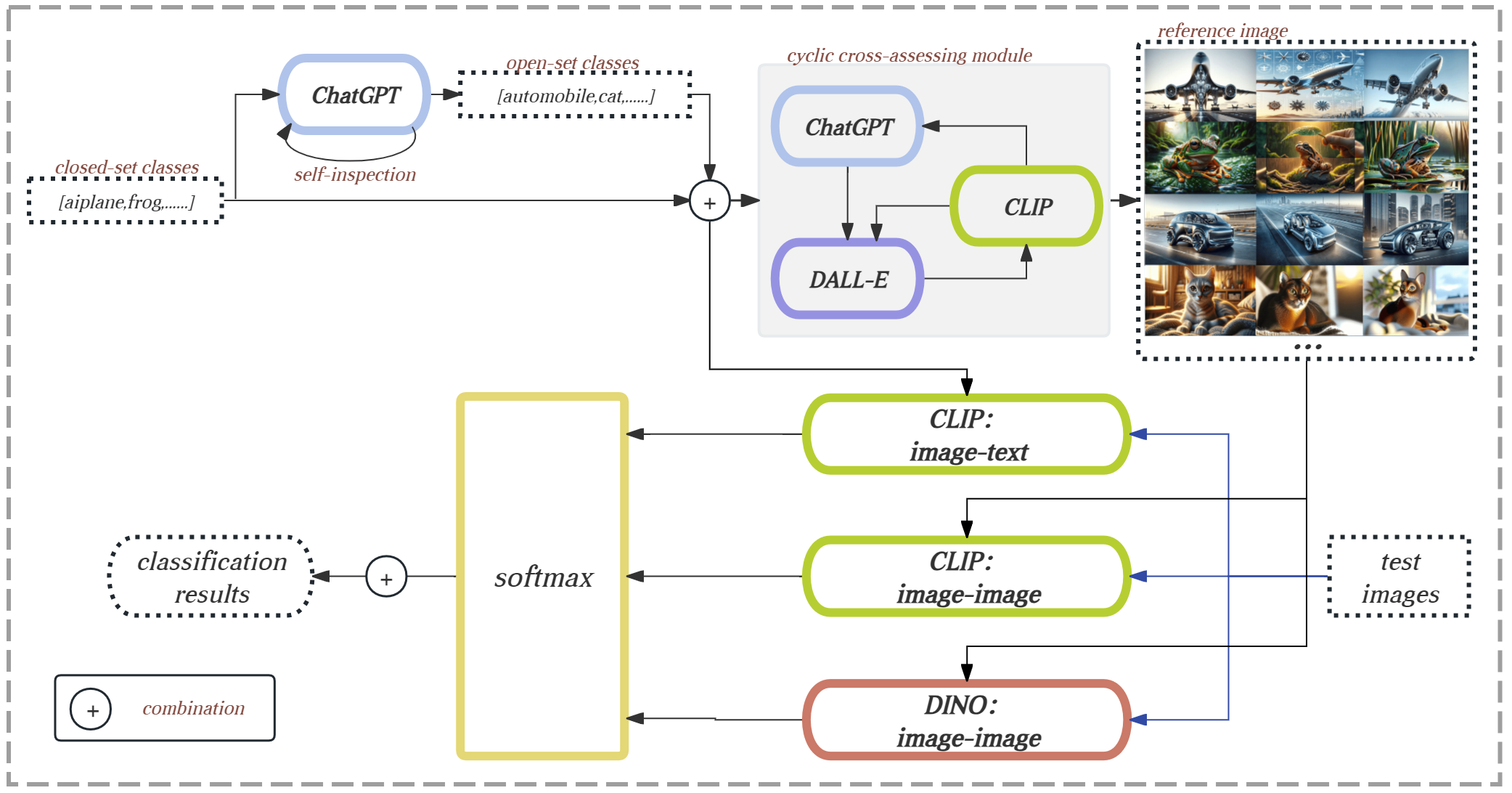}
    \caption{Overview of our method.}
    \label{main_fig}
\end{figure*}

\section{Method}
The overview of our method is summarized in Figure \ref{main_fig}. Assume we have a dataset defined as $D$, containing $m$ seen classes and $n$ unseen classes. When implementing the classification task of $m+n$ (defined as $N$) categories within the dataset $D$, we first generate some reference images for each category. After that, for a given test image $x_{t}$, we first align $x_{t}$ with class-definition text using the CLIP model's text-image alignment technique, followed by image-image alignment with the CLIP model and image-image alignment with the DINO model that utilize the previously-generated images as reference. Finally, confidence levels are used as weights to combine the three methods, yielding the final result. In the subsequent Section \ref{text_clip}, we briefly introduce the image classification method of CLIP. In Section \ref{text_generate}, we discuss how to generate reference images, and in Section \ref{text_prediction}, we explain how to use these generated reference images to achieve image-image alignment with CLIP and DINO. Finally, Section \ref{prediction_fusion} illustrates how to integrate the three methods to produce the final result. 
\par We utilize the extensive knowledge of ChatGPT and the powerful image generation capabilities of DALLE to create reference images that precisely describe unseen categories and classification boundaries, obtaining reference samples for new categories and those at the classification boundaries. This differs from the approach by  \cite{82}, which aimed to augment training samples using DALLE. Unlike the preprocessing done during training in previous methods, our approach supplements high-quality boundary images during prediction as reference images. Additionally, we propose leveraging CLIP's image feature extraction capability, integrating not only its text-image alignment but also image-image alignment techniques during classification. Compared to models without this technique, ours achieved a respective increase of 0.45\%, 0.24\%, and 0.23\% in classification accuracy on the CIFAR10, CIFAR100, and TinyImageNet datasets. Furthermore, we introduce a confidence-based adaptive weighting mechanism to aggregate results from different prediction models. This mechanism allows us to weigh the results of each prediction model based on its confidence level, enabling more effective integration of multiple model predictions. Such adaptive weighting enhances the flexibility of our approach to accommodate varying performances of different models in diverse scenarios, further improving overall prediction accuracy and robustness.
Regarding the architecture of CLIP, it comprises two main components: an image encoder ($F_{i}$) and a text encoder ($F_{t}$). The image encoder takes an image $I$ as input and extracts its feature representation through a series of transformer blocks; and the text encoder takes text $T$ as input and extracts its feature representation through another set of transformer blocks. Formally,
\begin{equation}
    f_{i} = F_{i}\left(I\right);\; f_{t} = F_{t}\left(T\right).
\end{equation}
Through pretraining, the outputs of both image and text encoders have been mapped into a shared representational space. By leveraging this property, we can implement zero-shot classification tasks by using cosine similarity to calculate the alignment of a given test image $x_{t}$ and a text $T_{c}$ that represents the class $c$, i.e, `\texttt{a photo of [class-c]}'. Formally, this process is formulated as:
\begin{equation}
    S = \frac{F_{i}(I)\cdot F_{t}(T_{c})}{||F_{i}(I)||\ ||F_{t}(T_{c})||}.
\end{equation}
This result reflects the degree of similarity between the text and image representations in the feature space. When classifying images, we can align the image features with $N$ such text features for all $N$ classes and then select the class with the highest similarity as the classified category. In this way, the classification can be implemented. 

\begin{figure*}[t]
    \centering
    \includegraphics[width=1.0\linewidth]{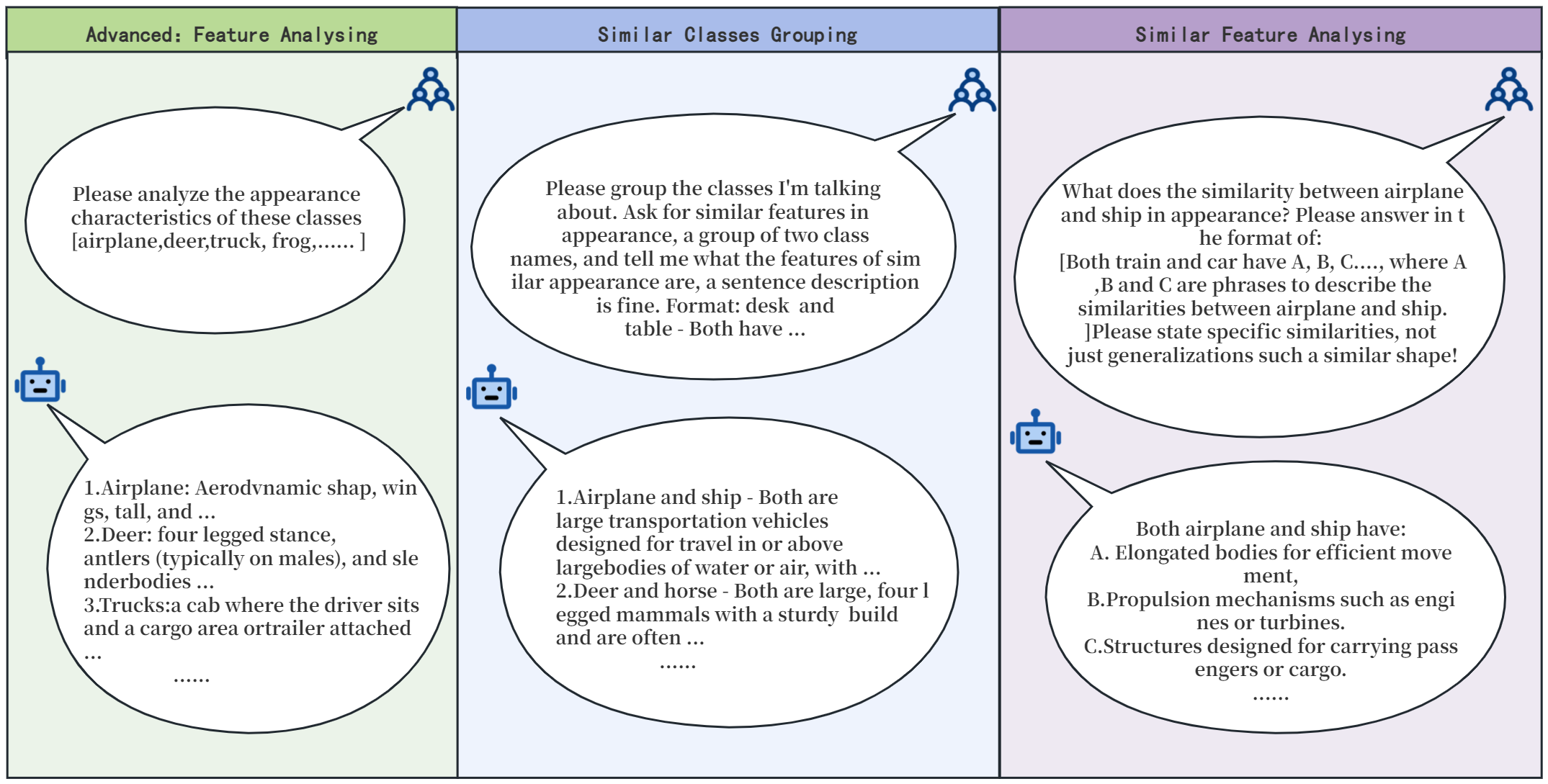}
    \caption{Examples of utilizing ChatGPT to generate common features of similar categories.}
    \label{chat_fig}
\end{figure*}

\subsection{CLIP Model} \label{text_clip}
We first provide an introduction to the CLIP model. CLIP, by pre-training simultaneously on images and text, learns to map them into a shared representational space where semantically related images and texts have similar features in this space.
\subsection{Reference Images Generation} \label{text_generate}
Although the above-mentioned method can achieve a relatively simple zero-shot classification by using CLIP-based image-text alignment, its performance is still not satisfactory enough due to the cross-domain mismatching problems. This is because texts and images belong to different domains, so strictly aligning them to the same space is still very challenging, especially when dealing with categories that are unseen or are long-tailed with very few samples in training. To alleviate this issue, we propose a novel approach that further introduces image-image alignment in addition to the original image-text alignment, which helps to produce a more precise prediction by preventing the cross-domain mismatching problem. 

However, directly employing image-image alignment is not feasible in our task setting, as there are no images available for the unseen novel classes. To solve this problem, we propose to fully leverage the rich knowledge of ChatGPT and the powerful image generation capabilities of DALL-E to  synthesize images of these categories to serve as references for image-image alignment. Note that many categories are easily confused in appearance with other categories since they may be similar in shape or features. Thus to achieve the clearer distinguishment between these categories, inspired by SVM, we propose to generate reference images that are located near the classification boundaries of the ambiguous categories. To be specific, the references images are produced through the following procedures:

First, we utilize ChatGPT to briefly analyze the appearance characteristics of these $N$ categories. The purpose of this step is to familiarize ChatGPT with the characteristics of all categories, thus enabling it to better perform subsequent grouping and reference generation operations. The prompt for inputting into ChatGPT in this step is: `Please analyze the appearance characteristics of these classes [`goldfish', `bullfrog', `tailed\_frog',... ...]'. Having analyzed the appearance features of all categories, ChatGPT can more accurately and comprehensively select similar combinations. 

The next step is to use ChatGPT to group categories that are similar in appearance, which helps us to obtain a clearer classification boundary in the further steps. Specifically, we ask ChatGPT to identify those similar categories that are easily confused, which is enabled by inputting ChatGPT with the following prompt: `Please group the classes I'm talking about without using any other class names. Ask for similar features in appearance, a group of two class names, and tell me what the features of similar appearance are, a sentence description is fine. Format: desk and dining\_table - Both have flat horizontal surfaces with legs for support'. To ensure the comprehensiveness and accuracy of the generated combinations, we confirm with ChatGPT again: `Is there any other combination with similar appearance? If not, answer no'. Through these methods, we can obtain groups of similar combinations. 

Then, for each pair of combinations given in ChatGPT's response, we further analyze the common appearance features of each category pair. To achieve this, taking the identified combination  `train’ and `car’ as an example, we use the following prompt for inputting into ChatGPT: `What does the similarity between train and car in appearance? Please answer in the format of: both train and car have A, B, C...., where A, B, and C are phrases to describe the similarities between train and car. Please state specific similarities, not just generalizations such as similar shape!” In this way, we get the common features generated by GPT and denote it as $cc$. 

Lastly, we use DALL-E to generate images of each class that possess the common appearance features $cc$ as its confused category, with the prompt: `generate an image of [category name] that has [common characteristics cc]. As realistic as possible. More fit for life.'. It is noteworthy that there exist many categories that do not resemble other categories. For such categories, we directly utilize DALL-E to generate images belonging to it without using detailed appearance descriptions. To acquire more comprehensive reference information, we generate multiple images for each category. All these images will be employed in the subsequent steps for image-image alignment.

\subsection{Classification Prediction} \label{text_prediction}
As discussed before, aligning images and texts using only CLIP is insufficient. Thus in the previous section, we generate reference images for further helping to implement image-image alignment. Additionally, we find that another zero-shot classifier, DINO, has also achieved tremendous success in the field of computer vision. Therefore, we propose to utilize both the CLIP and DINO to perform zero-shot classification as follows:

First, we perform the original CLIP-based image-text alignment. To be specific, given a test image $x_{t}$, we employ CLIP's image encoder $F_{i}$ to extract its feature $F_{i}(x_{t})$ and compare it with the text feature $F_{t}(T_{n})$ for the $n$-th class, where $T_{n}$ denotes a text corresponding to the $n$-th category and its format is `\texttt{A photo of [class-n]}'. Formally,
\begin{equation}
    S_{i-t, clip}^{n} = {\rm Cos}\left(F_{i}(x_{t}), F_{t}(T_{n})\right).
\end{equation}
In this way, we get $S_{i-t, clip}\in \mathbb{R}^{N}$ for all $N$ categories. 

Next, we perform image-image alignment based on the CLIP model. Concretely, denote the reference images for the $n$-th category as $\{r_{n}^{i}\}_{i=1}^{M}$, where $M$ is the number of reference images for each category. Then the alignment in this step is calculated as:
\begin{equation}
    S_{i-i, clip}^{n} = \frac{1}{M} \sum_{i=1}^{M}{\rm Cos}\left(F_{i}(x_{t}), F_{i}(r_{n}^{i})\right).
\end{equation}
In this way, we get $S_{i-i, clip}\in \mathbb{R}^{N}$ for all $N$ categories. 

Finally, we use the same method to get the image-image alignment result $S_{i-i, clip}$ based on the DINO model:
\begin{equation}
    S_{i-i, dino}^{n} = \frac{1}{M} \sum_{i=1}^{M}{\rm Cos}\left(D(x_{t}), D(r_{n}^{i})\right).
\end{equation}
where $D$ refers to the image encoder of DINO. 

\subsection{Prediction Fusion} \label{prediction_fusion}
The three types of prediction results obtained in the aforementioned manner are subsequently fused to produce the final classification outcome. To enhance the reliability of the fused results, we propose a confidence-based fusion mechanism that assigns greater weight to predictions with higher confidence during the fusion process. Concretely, denote $S$ as the prediction result for one of the above-introduced three alignment approaches, we propose the following three methods to calculate its confidence level $W$:
(1) using the maximum value on each $S$ as the confidence level: 
\begin{equation}
    W = {\rm max}\{S^{n}\}_{n=1}^{N}.
\end{equation}

(2) using the inverse of the model's entropy as the confidence value:
\begin{equation}
    H = -\sum_{n}^{N} S^{n}\log(S^{n}),\ W = \frac{1}{H+1e-6}.
\end{equation}
where $1e-6$ is a small constant that avoids having a denominator of 0.

(3)  using the negative exponential of the model's entropy as the confidence value:
\begin{equation}
    H = -\sum_{n}^{N} S^{n}\log(S^{n}),\ W = {\rm exp}\left(-H\right).
\end{equation}

In the experimental section, we'll assess to determine the optimal method yielding the highest results. Subsequently, we utilize the obtained confidences from $\{S_{i-t,clip}, S_{i-i,clip}, S_{i-i, dino}\}$ as fusion weights for these outcomes.

\begin{equation}
    S_{fuse} = W_{i-t,clip}\cdot S_{i-t,clip} + W_{i-i,clip}\cdot S_{i-i,clip} + W_{i-i, dino}\cdot S_{i-i, dino}.
\end{equation}

\section{Experiments}
\subsection{Datasets and Evaluation Metrics}
To evaluate the effectiveness of our proposed method, we conduct experiments on three widely-used datasets, including CIFAR10, CIFAR100, and TinyImageNet. For the CIFAR10 dataset that contains 10 categories, we select 6 categories as the closed set and the remaining 4 categories as the open set. For the CIFAR10 dataset that contains 100 categories, 10 categories are chosen as the closed set, with the remaining 90 categories serving as the open set. For the TinyImageNet dataset: we select 20 categories as the closed set, with the other 180 categories as the open set. We employ two metrics to measure model performance: Accuracy (AC) and Area Under the Receiver Operating Characteristic curve (AUROC). The accuracy metric is divided into top-1, top-3, and top-5, each reflecting the frequency of correct categories appearing in the model's predictions for the most likely one, three, and five categories, respectively. This allows for a comprehensive assessment of the model's performance in closed-set classification tasks. The AUROC metric is used to evaluate the model's ability to differentiate between closed-set and open-set samples, that is, the accuracy of the model in identifying novel category samples.

\begin{table}[!h]
\centering
\caption{Top1, Top3, Top5 results for single method and and our three-method fusion.}
\label{tab:classification_accuracy}
\renewcommand{\arraystretch}{1.2} 
\scalebox{0.76}{%
\begin{tabular}{@{}lccccccccc@{}}
\toprule
Method & \multicolumn{3}{c}{CIFAR10} & \multicolumn{3}{c}{CIFAR100} & \multicolumn{3}{c}{TinyImageNet} \\ 
\cmidrule(r){2-4} \cmidrule(lr){5-7} \cmidrule(l){8-10}
                          & Top1  & Top3  & Top5  & Top1  & Top3  & Top5  & Top1  & Top3  & Top5  \\ \midrule
CLIP text-image (M1)         & 79.99\% & 93.22\% & 97.41\% & 47.69\% & 66.14\% & 72.35\% & 41.20\% & 61.16\% & 69.24\% \\
CLIP image-image (M2)         & 65.90\% & 86.20\% & 86.20\% & 30.43\% & 48.55\% & 56.67\% & 31.56\% & 48.17\% & 55.57\% \\
DINO image-image (M3)         & 79.36\% & 90.48\% & 94.51\% & 66.75\% & 81.77\% & 86.48\% & 72.32\% & 72.32\% & \textbf{86.79\%} \\
ours (M1+M2+M3)           & \textbf{91.96\%} & \textbf{98.14\%} & \textbf{99.30\%} & \textbf{72.17\%} & \textbf{86.55\%} & \textbf{90.36\%} & \textbf{73.52\%} & \textbf{84.95\%} & 88.55\% \\ \bottomrule
\end{tabular}}
\end{table}

\begin{table}[!h]
\centering
\caption{AUROC results for single method and our three-method fusion.}
\label{tab:auroc_results}
\begin{tabular}{lccc}
\toprule
Method & CIFAR10 & CIFAR100 & TinyImageNet \\
\midrule
CLIP text-image (M1) & 97.26\% & 91.41\% & 85.95\% \\
CLIP image-image (M2) & 98.12\% & 85.73\% & 82.68\% \\
DINO image-image (M3) & 97.97\% & 94.28\% & 96.07\% \\
Ours (M1+M2+M3) & \textbf{99.78\%} & \textbf{96.03\%} & \textbf{96.48\%} \\
\bottomrule
\end{tabular}
\end{table}

\subsection{Main Results}
As discussed detailedly in the method section, the final result of our method is based on the fusion of three single approaches: image-text alignment based on CLIP (M1), image-image alignment based on CLIP (M2) and image-image alignment based on DINO (M3). The CLIP model, as an innovative learning framework, enhances cross-modal recognition accuracy by bridging text and image information. Meanwhile, the approach based on DINO utilizes self-supervised learning mechanisms to strengthen the extraction and recognition capabilities of image features. In Table \ref{tab:classification_accuracy} and Table \ref{tab:auroc_results}, we compare the performance of our proposed fusion method with these single alignment approaches. The results show that our method can achieve significantly better performance than all other single methods in terms of all metrics including TOP-1, TOP-3, TOP-5 and AUROC. These results demonstrate that by integrating the advantages of different models and alignment manners, our method can process complex data more effectively with higher generalization ability. 

\subsection{Analysis of Main Results}
Through the experimental results mentioned above, we discover that although each method has its unique advantages, they also face a series of challenges when applied independently. An analysis of the experimental results for these single-alignment methods allows us to understand the challenges faced by each model. 

Specifically, the CLIP image-text alignment method (M1) perform well on the CIFAR10 dataset but is outperformed by other methods on the CIFAR100 and TinyImageNet datasets. This may be because the CLIP model is primarily designed for text and image alignment, and its performance might be constrained by the complexity and diversity of image data. On complex datasets, textual information may not provide sufficient assistance, leading to a decrease in performance.

The CLIP image-image alignment method (M2) underperfors M1 across all datasets, particularly on the CIFAR100 and TinyImageNet datasets. Since the CLIP model is mainly designed for text and image alignment, it might face challenges in aligning images with each other. It may not effectively capture the similarity between images, resulting in decreased performance.

The DINO image-image alignment method (M3) outperforms M1 and M2 on all datasets but still fell short of the fused model's performance. Although the DINO model leverages self-supervised learning mechanisms, its performance on certain datasets might still be limited. There could be inaccuracies in image feature extraction or a lack of sensitivity to the alignment between images.

The performance differences among individual methods indicate that each method may have its strengths and weaknesses on different datasets. Therefore, by combining the text and image alignment capabilities of the CLIP model with the image feature extraction abilities of the DINO model, we can enhance the performance and generalization capabilities of different models and alignment methods, thereby achieving better results.

In addition to the challenges faced by individual models, our research also encounters other challenges, especially when dealing with cross-domain issues and limitations related to the reference image set:

(1) Cross-domain issues: Single models typically perform well within specific domains, but their performance may be limited when dealing with cross-domain or diverse data. For instance, the text-image alignment based on CLIP (M1) can effectively handle the correspondence between text descriptions and image content. However, its performance may decline in certain situations, such as when image content is complex or text descriptions are vague. Similarly, the image-image alignment approaches based on CLIP and DINO (M2 and M3), while capable of capturing visual features, may struggle to effectively transfer knowledge when faced with images from different domains (such as artwork versus natural scenes).

(2) Limitations of reference image generation: The generation and selection of reference images are crucial for alignment-based models. In our setup, the scale and diversity of the reference image set might be limited, meaning the model may not cover all possible categories or scenes. For CLIP's image-image alignment (M2) and DINO's image-image alignment (M3), if the reference images are insufficient to represent the diversity in the test images, the model's discriminative ability will be limited. Additionally, a small set of reference images may also lead to poor model performance when confronted with unseen categories or edge cases.

Through these insights, we have gained a deeper understanding of the contributions of different model fusion strategies to enhancing overall performance, as well as the relative strengths and limitations of various models on specific tasks. These findings provide important references for future research in the field of computer vision, especially in dealing with cross-modal data fusion and enhancing model robustness.

\subsection{Ablation Study}
\subsubsection{Effectiveness of Different Components}
To verify the effectiveness for different components of our method, we test the performance of different combinations and present the results Table \ref{tab:classification_accuracy_2_3_model_fusion} and Table \ref{tab:auroc_2_3_model_fusion}. According to results, we find that by removing any part from our method, the obtained combinations of M1 and M2, M1 and M3, or M2 and M3 would have a decrease in performance compared to our full method (M1+M2+M3). These results demonstrate the importance of each component in our method.  

\begin{table}[!h]
\centering
\caption{Top1, Top3, Top5 results for two-method fusion and our three-method fusion.}
\label{tab:classification_accuracy_2_3_model_fusion}
\renewcommand{\arraystretch}{1.4} 
\scalebox{0.76}{
\begin{tabular}{lccccccccc}
\toprule
Method  & \multicolumn{3}{c}{CIFAR10} & \multicolumn{3}{c}{CIFAR100} & \multicolumn{3}{c}{TinyImageNet} \\
\cmidrule(r){2-4} \cmidrule(lr){5-7} \cmidrule(l){8-10}
                         & Top1     & Top3     & Top5     & Top1     & Top3     & Top5     & Top1     & Top3     & Top5     \\
\midrule
M1+M2                    & 83.83\%  & 94.85\%  & 98.22\%  & 48.00\%  & 66.41\%  & 73.27\%  & 43.52\%  & 63.97\%  & 71.79\%  \\
M1+M3                    & 92.51\%  & 98.16\%  & \textbf{99.35\%}  & 71.93\%  & \textbf{86.59\%}  & \textbf{90.50\%}  & 73.29\%  & 84.73\%  & 88.41\%  \\
M2+M3                    & 79.99\%  & 93.22\%  & 97.41\%  & 64.19\%  & 79.29\%  & 84.63\%  & 72.74\%  & 84.33\%  & 87.30\%  \\
Ours (M1+M2+M3)          & \textbf{92.96\%}  & \textbf{98.32\%}  & 99.33\%  & \textbf{72.17\%}  & 86.55\%  & 90.36\%  & \textbf{73.52\%}  & \textbf{84.95\%}  & \textbf{88.55\%}  \\
\bottomrule
\end{tabular}}
\end{table}

\begin{table}[!h]
\centering
\caption{AUROC results for two-method fusion and our three-method fusion.}
\label{tab:auroc_2_3_model_fusion}
\begin{tabular}{lccc}
\toprule
Method & CIFAR10 & CIFAR100 & TinyImageNet \\
\midrule
M1+M2 & 97.94\% & 92.99\% & 87.03\% \\
M1+M3 & 99.75\% & 95.87\% & \textbf{96.54\%} \\
M2+M3 & 97.26\% & 94.70\% & 96.09\% \\
Ours (M1+M2+M3) & \textbf{99.78\%} & \textbf{96.03\%} & 96.48\% \\
\bottomrule
\end{tabular}
\end{table}

We further provide an in-depth analysis of these results. Concretely, we observe that on some datasets, the combination of M1 and M3 sometimes perform better in terms of top3 and top5 prediction results than the fusion of all three methods. However, this difference is not significant and does not clearly surpass the overall effect of our three-method fusion. One possible reason for this might be the relatively small size of the test dataset, which consists of only ten thousand images. This could lead to certain combination methods showing slight advantages on specific metrics, warranting further adjustments and modifications in future studies. Notably, across all datasets, the three-method fusion approach consistently performs best on the Top1 metric, clearly outperforming any other two-method combinations. This indicates that in most cases, the effect of fusing three method leads in performance over other combination methods. Overall, the three-method fusion approach demonstrates significant advantages on different datasets and across various metrics in the vast majority of cases, further validating the effectiveness and stability of the comprehensive method we propose.

\begin{table}[ht]
\centering
\caption{Top1, Top3, Top5 results for different weight settings on the CIFAR10 Dataset}
\label{tab:weight_setting_methods}
\begin{tabular}{lccc}
\toprule
Weighting Method& Top1 & Top3 & Top5 \\
\midrule
1:1:1 & 92.36\% & 98.22\% & 99.31\% \\
3:3:4 & 92.29\% & 98.21\% & 99.26\% \\
Max Similarity & 92.75\% & 98.26\% & 99.34\% \\
Inverse Entropy & \textbf{92.96\%} & \textbf{98.32\%} & 99.33\% \\
Negative Exponential of Entropy & 92.90\% & 98.31\% & \textbf{99.36\%} \\
\bottomrule
\end{tabular}
\end{table}

\begin{table}[!h]
\centering
\caption{AUROC results for different weight setting methods on the CIFAR10 dataset.}
\label{tab:auroc_weight_setting_methods}
\begin{tabular}{lc}
\toprule
Weighting Method & AUROC \\
\midrule
1:1:1 & 99.55\% \\
3:3:4 & 99.60\% \\
Max Similarity & 99.68\% \\
Inverse Entropy & \textbf{99.78\%} \\
Negative Exponential of Entropy & 99.73\% \\
\bottomrule
\end{tabular}
\end{table}

\subsubsection{Effectiveness of Different Fusion Methods}
In the context of multi-model fusion, a common challenge is how to effectively fuse information from different sources to enhance the final decision-making or classification performance. This work proposes an entropy-based weighting strategy for fusing similarity results from different models (CLIP and DINO) and different alignment manners (image-text alignment and image-image alignment). As discussed in Section \ref{prediction_fusion}, we propose three different methods to calculate the confidence levels that serve as the weights for the further fusion. Here, we conduct experiments on CIFAR10 to find the optimal method from these candidates. Note that in addition to these proposed methods that dynamically fuse different results based on their confidence levels, we also evaluate a straightforward way that manually sets the weighting coefficient as 1:1:1 and 3:3:4 for $W_{i_t, clip}$, $W_{i-i, clip}$ and $W_{i-i, dino}$, respectively. The results in terms of top1, top3, top5 accuracy rates are presented in Table \ref{tab:weight_setting_methods}, and the results of AUROC metric are presented in Table \ref{tab:auroc_weight_setting_methods}. Upon comparing the performance of different methods, we observe that the method utilizing the  `reciprocal of information entropy' consistently shows the best performance. Looking at the overall data, on the CIFAR10 dataset, the entropy-based reciprocal weighting method we proposed achieved the most optimal prediction performance. This result strongly supports our research hypothesis, confirming the effectiveness and superiority of this method on specific datasets. Therefore, we choose it as our fusion weighting method. 

\begin{table}[!h]
\centering
\caption{Comparison of Top1 and AUROC results when using one and multiple reference images. The left side of each `-' represents the Top1 result, and the right side of each `-' represents the AUROC result.}
\label{tab:one_multiple_reference_images}
\scalebox{0.9}{
\begin{tabular}{lccc}
\toprule
Number of Reference Images & CIFAR10 & CIFAR100 & TinyImageNet \\
\midrule
One & 88.71\%-99.05\% & 67.78\%-96.025\% & 67.77\%-96.00\% \\
Multiple & \textbf{92.96\%-99.78\%} & \textbf{72.17\%-96.026\%} & \textbf{73.52\%-96.48\%} \\
\bottomrule
\end{tabular}}
\end{table}

\begin{table}[htbp]
\centering
\caption{The AUROC results on the detection of closed-set and open-set samples. Results are averaged over multiple random dataset splits following .}
\scalebox{0.96}{
\begin{tabular}{lllll}
\toprule
Method & Venue & CIFAR10 & CIFAR+10 & TinyImageNet \\
\midrule
\multicolumn{5}{c}{\textbf{Methods that involve a training process}} \\
\midrule
RPL\cite{71} & ECCV 2020 & $90.1 \pm -$ & $97.6 \pm -$ & $80.9 \pm -$ \\
OpenHybrid\cite{72}  & ECCV 2020 & $95.0 \pm -$ & $96.2 \pm -$ & $79.3 \pm -$ \\
PMAL\cite{75}  & AAAI 2022 & $95.1 \pm -$ & $97.8 \pm -$ & $83.1 \pm -$ \\
ZOC\cite{76}  & AAAI 2022 & $93.0 \pm 1.7$ & $97.8 \pm 0.6$ & $84.6 \pm 1.0$ \\
MLS\cite{77}  & ICLR 2022 & $93.6 \pm -$ & $97.9 \pm -$ & $83.0 \pm -$ \\
Class-inclusion\cite{79}  & ECCV 2022 & $94.8 \pm -$ & $96.1 \pm -$ & $78.5 \pm -$ \\
CSSR\cite{81}  & TPAMI 2022 & $91.3 \pm -$ & $96.3 \pm -$ & $82.3 \pm -$ \\
RCSSR\cite{81}  & TPAMI 2022 & $91.5 \pm -$ & $96.0 \pm -$ & $81.9 \pm -$ \\
\midrule
\multicolumn{5}{c}{\textbf{Methods that involve no extra training process}} \\
\midrule
Ours & & \textbf{99.78} & 96.03 & \textbf{96.48} \\
\bottomrule
\end{tabular}}
\end{table}

\subsubsection{The Number of Reference Images}
In our method, the number of generated reference images can has a impact on the ability to distinguish between different categories. To explore this further, we conduct experiments to compare the method that uses only one reference image with that uses multiple reference images per category. As shown in Table \ref{tab:one_multiple_reference_images}, the results indicate that using multiple reference images outperforms using one reference image significantly. By using more reference images, we can obtain a more general information that can better represent a category's properties, which can serve as a more effective guidance to help perform classification more accurately.

\begin{figure*}[t]
    \centering
    \includegraphics[width=1.0\linewidth]{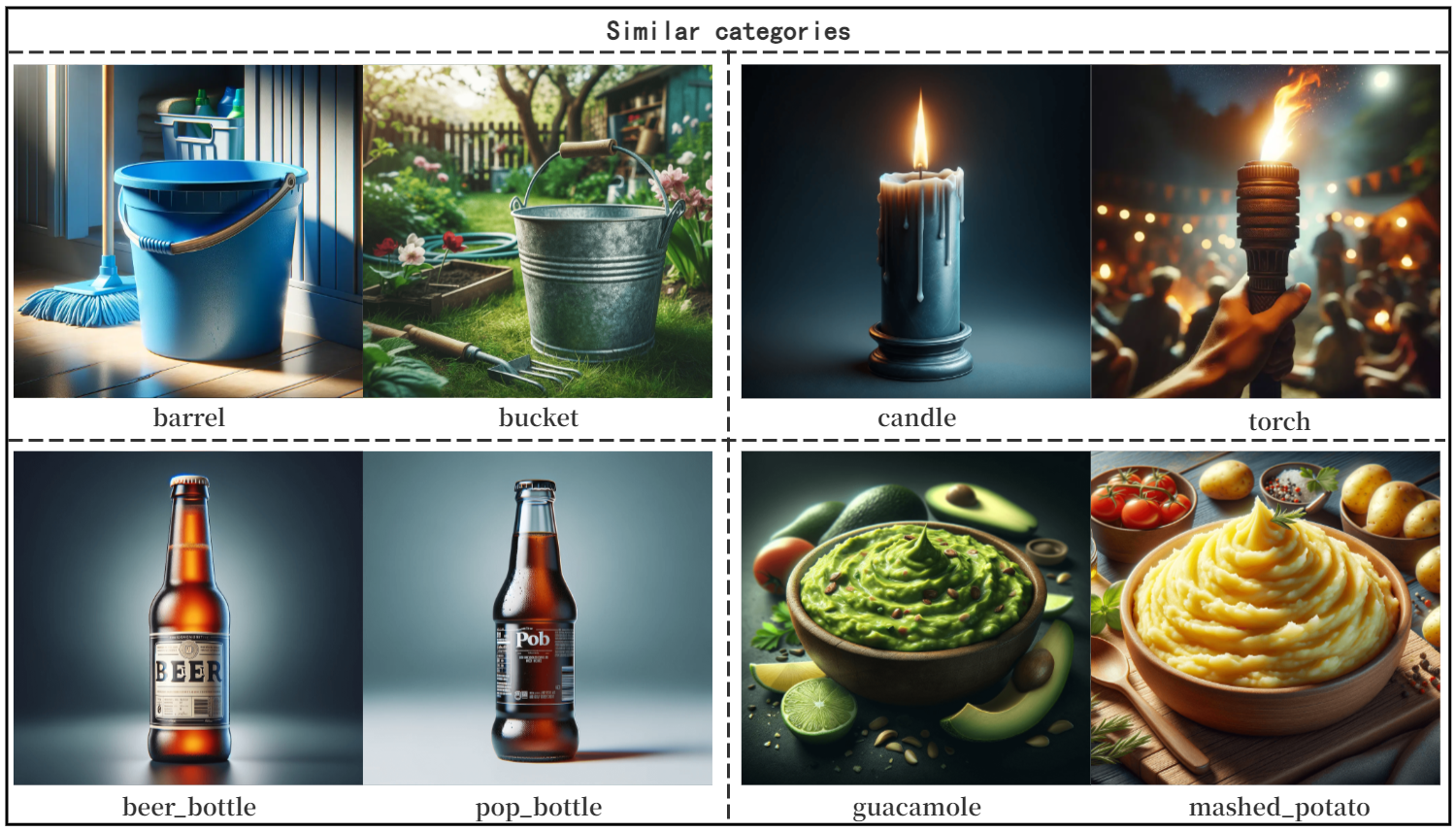}
    \caption{Examples of synthesized reference images for similar categories with similar appearance. }
    \label{example_ref_image}
\end{figure*}

\subsection{Visualizations}
 As introduced in Section \ref{text_generate}, we propose to generate reference images for unseen new classes. These reference images will then be used for the image-image alignment in the further processes. To learn clearer classification boundaries for enhancing accuracy, we make the synthesized reference images to have the similar appearance as their easily confusable similar categories. Some examples of these synthesized reference images are presented in Figure \ref{example_ref_image}. 

\begin{figure*}[t]
    \centering
    \includegraphics[width=1.0\linewidth]{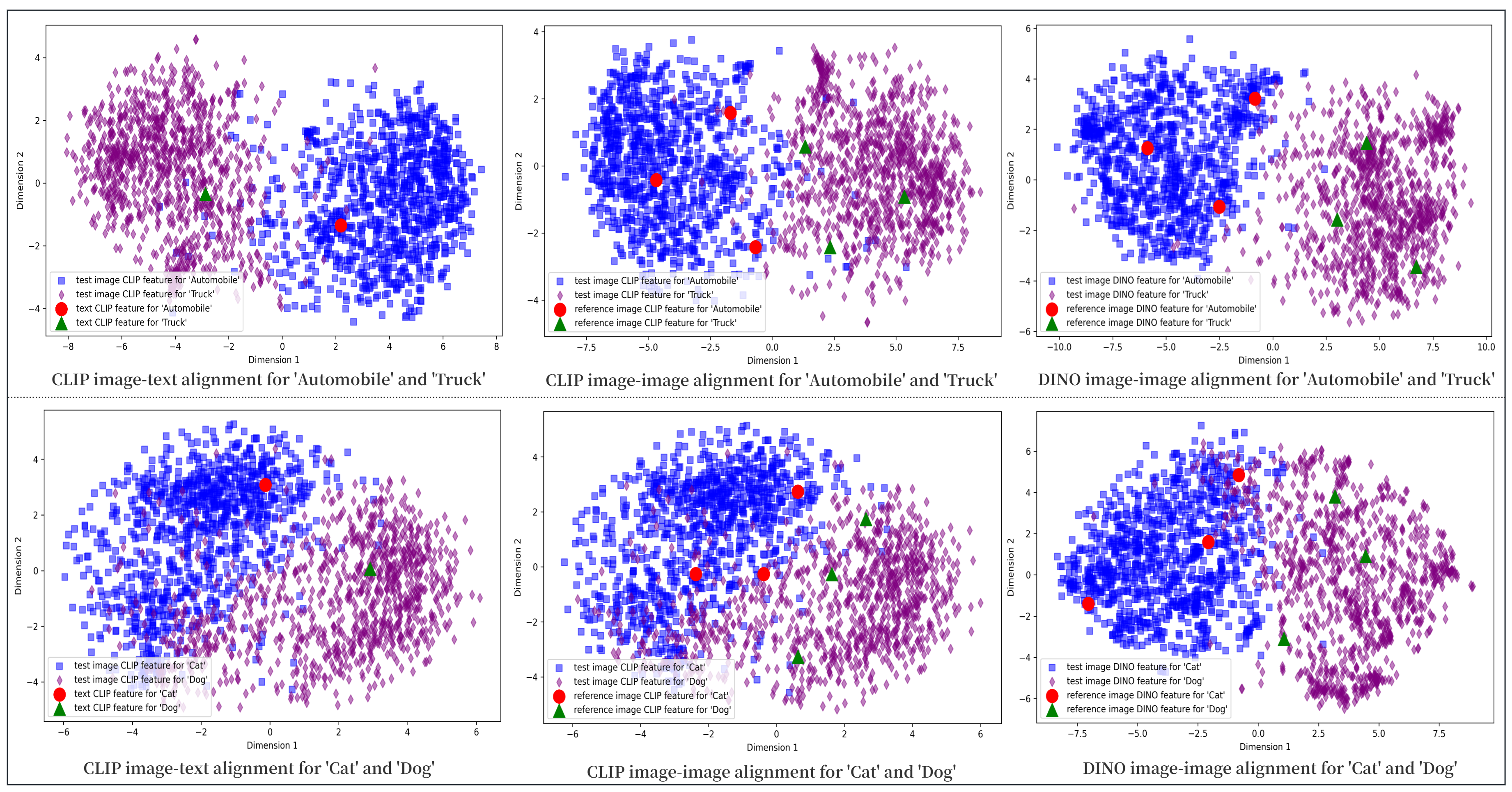}
    \caption{Examples of t-SNE results for CLIP image-text alignment, CLIP image-image alignment and DINO image-image alignment. }
    \label{tsne}
\end{figure*}

We further conduct a detailed analysis for the features of two similar groups of categories [cats, dogs] and [cars, trucks] within the CIFAR-10 dataset. Concretely, for each category, we visualize the features of all test images that belongs to the category, and the features of its reference images along with the text features of `A photo of [class]'. We employ the t-SNE technique for visualizations and the results for CLIP image-text alignment, CLIP image-image alignment an DINO image-image alignment are presented in Figure \ref{tsne}. As an efficient dimensional reduction algorithm, t-SNE allows us to intuitively display and compare the data distribution and category boundaries of different models in high-dimensional feature spaces.

More concretely, we pay special attention to the generated reference images, which play a key role in discriminating classification boundaries for our final prediction. Through the visualization results shown in Figure \ref{tsne}, we observe that feature points of many reference images are located at the boundaries between categories, which can help us to distinguish similar categories more clearly. Furthermore, we observe that feature points of different categories in the DINO model show higher clustering compared to in the CLIP model, likely due to its better capture of classification decision boundaries during training.

\section{Conclusion}
In this paper, we introduce an innovative image classification framework designed to tackle the challenges encountered by traditional techniques when facing unknown categories. Our approach is founded on three core strategies: 1) Utilizing the capabilities of ChatGPT and DALL-E, we generate reference images that accurately describe new categories, effectively addressing the issue of insufficient information. 2) By integrating the technologies of CLIP and DINO, we achieve a deep alignment between text and images, enhancing the accuracy of predictions. 3) We have introduced a mechanism that adjusts weights based on confidence levels, optimizing the amalgamation of different model predictions. Through these three innovative strategies, our method demonstrates significant performance advantages in dealing with unseen categories. Experimental results on datasets such as CIFAR-10, CIFAR-100, and TinyImageNet validate the effectiveness of our approach, showcasing its exceptional performance in both open and closed set classification tasks.


\begin{thebibliography}{10}

\bibitem{55}
K.~He, X.~Zhang, S.~Ren, and J.~Sun.
\newblock Deep residual learning for image recognition.
\newblock In {\em Proceedings of the IEEE conference on computer vision and pattern recognition}, pages 770--778, 2016.

\bibitem{57}
A.~G. Howard, M.~Zhu, B.~Chen, D.~Kalenichenko, W.~Wang, T.~Weyand, M.~Andreetto, and H.~Adam.
\newblock Mobilenets: Efficient convolutional neural networks for mobile vision applications.
\newblock {\em arXiv preprint arXiv:1704.04861}, 2017.

\bibitem{54}
T.~Paz-Argaman, Y.~Atzmon, G.~Chechik, and R.~Tsarfaty.
\newblock Zest: Zero-shot learning from text descriptions using textual similarity and visual summarization.
\newblock {\em arXiv preprint arXiv:2010.03276}, 2020.

\bibitem{1}
T.~Brown, B.~Mann, N.~Ryder, M.~Subbiah, J.~Kaplan, P.~Dhariwal, A.~Neelakantan, P.~Shyam, G.~Sastry, A.~Askell, et~al.
\newblock Language models are few-shot learners.
\newblock {\em Advances in neural information processing systems}, 33:1877--1901, 2020.

\bibitem{2}
J.~Devlin, M.~W. Chang, K.~Lee, and K.~Toutanova.
\newblock Bert: Pre-training of deep bidirectional transformers for language understanding.
\newblock {\em arXiv preprint arXiv:1810.04805}, 2018.

\bibitem{3}
G.~Ou, G.~Yu, C.~Domeniconi, X.~Lu, and X.~Zhang.
\newblock Multi-label zero-shot learning with graph convolutional networks.
\newblock {\em Neural Networks}, 132:333--341, 2020.

\bibitem{4}
J.~Gao and C.~S. Xu.
\newblock Ci-gnn: Building a category-instance graph for zero-shot video classification.
\newblock {\em IEEE Transactions on Multimedia}, 22(12):3088--3100, 2020.

\bibitem{8}
S.~Sankaranarayanan and Y.~Balaji.
\newblock Meta learning for domain generalization.
\newblock In {\em Meta Learning With Medical Imaging and Health Informatics Applications}, pages 75--86. Elsevier, 2023.

\bibitem{9}
Y.~Q. Xian, T.~Lorenz, B.~Schiele, and Z.~Akata.
\newblock Feature generating networks for zero-shot learning.
\newblock In {\em Proceedings of the IEEE conference on computer vision and pattern recognition}, pages 5542--5551, 2018.

\bibitem{7}
S.~C. Liu, M.~S. Long, J.~M. Wang, and M.~I. Jordan.
\newblock Generalized zero-shot learning with deep calibration network.
\newblock {\em Advances in neural information processing systems}, 31, 2018.

\bibitem{11}
J.~W. Ren, C.~J. Yu, X.~Ma, H.~Y. Zhao, S.~Yi, et~al.
\newblock Balanced meta-softmax for long-tailed visual recognition.
\newblock {\em Advances in neural information processing systems}, 33:4175--4186, 2020.

\bibitem{13}
A.~Radford, J.~W. Kim, C.~Hallacy, A.~Ramesh, G.~Goh, S.~Agarwal, G.~Sastry, A.~Askell, P.~Mishkin, J.~Clark, et~al.
\newblock Learning transferable visual models from natural language supervision.
\newblock In {\em International conference on machine learning}, pages 8748--8763. PMLR, 2021.

\bibitem{14}
J.~Shipard, A.~Wiliem, K.~N. Thanh, W.~Xiang, and C.~Fookes.
\newblock Diversity is definitely needed: Improving model-agnostic zero-shot classification via stable diffusion.
\newblock In {\em Proceedings of the IEEE/CVF Conference on Computer Vision and Pattern Recognition}, pages 769--778, 2023.

\bibitem{25}
A.~Christensen, M.~Mancini, A.~Koepke, O.~Winther, and Z.~Akata.
\newblock Image-free classifier injection for zero-shot classification.
\newblock In {\em Proceedings of the IEEE/CVF International Conference on Computer Vision}, pages 19072--19081, 2023.

\bibitem{26}
Z.~Novack, J.~McAuley, Z.~C. Lipton, and S.~Garg.
\newblock Chils: Zero-shot image classification with hierarchical label sets.
\newblock In {\em International Conference on Machine Learning}, pages 26342--26362. PMLR, 2023.

\bibitem{15}
Y.~Zhou, Y.~Yang, Q.~Ying, Z.~Qian, and X.~Zhang.
\newblock Multimodal fake news detection via clip-guided learning.
\newblock In {\em 2023 IEEE International Conference on Multimedia and Expo (ICME)}, pages 2825--2830. IEEE, 2023.

\bibitem{29}
N.~K. Lahajal et~al.
\newblock Enhancing image retrieval: A comprehensive study on photo search using the clip mode.
\newblock {\em arXiv preprint arXiv:2401.13613}, 2024.

\bibitem{31}
M.~Hendriksen, M.~Bleeker, S.~Vakulenko, N.~van Noord, E.~Kuiper, and M.~de~Rijke.
\newblock Extending clip for category-to-image retrieval in e-commerce.
\newblock In {\em European Conference on Information Retrieval}, pages 289--303. Springer, 2022.

\bibitem{16}
S.~T. Wasim, M.~Naseer, S.~Khan, F.~S. Khan, and M.~Shah.
\newblock Vita-clip: Video and text adaptive clip via multimodal prompting.
\newblock In {\em Proceedings of the IEEE/CVF Conference on Computer Vision and Pattern Recognition}, pages 23034--23044, 2023.

\bibitem{32}
M.~U. Khattak, H.~Rasheed, M.~Maaz, S.~Khan, and F.~S. Khan.
\newblock Maple: Multi-modal prompt learning.
\newblock In {\em Proceedings of the IEEE/CVF Conference on Computer Vision and Pattern Recognition}, pages 19113--19122, 2023.

\bibitem{17}
M.~Caron, H.~Touvron, I.~Misra, H.~Jégou, J.~Mairal, P.~Bojanowski, and A.~Joulin.
\newblock Emerging properties in self-supervised vision transformers.
\newblock In {\em Proceedings of the IEEE/CVF international conference on computer vision}, pages 9650--9660, 2021.

\bibitem{19}
F.~Li, H.~Zhang, H.~Xu, S.~Liu, L.~Zhang, L.~M. Ni, and H.-Y. Shum.
\newblock Mask dino: Towards a unified transformer-based framework for object detection and segmentation.
\newblock In {\em Proceedings of the IEEE/CVF Conference on Computer Vision and Pattern Recognition}, pages 3041--3050, 2023.

\bibitem{36}
H.~Zhang, F.~Li, S.~Liu, L.~Zhang, H.~Su, J.~Zhu, L.~M. Ni, and H.-Y. Shum.
\newblock Dino: Detr with improved denoising anchor boxes for end-to-end object detection.
\newblock {\em arXiv preprint arXiv:2203.03605}, 2022.

\bibitem{20}
A.~Radford, K.~Narasimhan, T.~Salimans, I.~Sutskever, et~al.
\newblock Improving language understanding by generative pre-training.
\newblock 2018.

\bibitem{39}
J.~A. Baktash and M.~Dawodi.
\newblock Gpt-4: A review on advancements and opportunities in natural language processing.
\newblock {\em arXiv preprint arXiv:2305.03195}, 2023.

\bibitem{40}
M.~Alawida, S.~Mejri, A.~Mehmood, B.~Chikhaoui, and O.~Isaac~Abiodun.
\newblock A comprehensive study of chatgpt: Advancements, limitations and ethical considerations in natural language processing and cybersecurity.
\newblock {\em Information}, 14(8):462, 2023.

\bibitem{42}
C.~E. Haupt and M.~Marks.
\newblock Ai-generated medical advice—gpt and beyond.
\newblock {\em JAMA}, 329(16):1349--1350, 2023.

\bibitem{43}
C.~Wu, J.~Lei, Q.~Zheng, W.~Zhao, W.~Lin, X.~Zhang, X.~Zhou, Z.~Zhao, Y.~Zhang, Y.~Wang, et~al.
\newblock Can gpt-4v (ision) serve medical applications? case studies on gpt-4v for multimodal medical diagnosis.
\newblock {\em arXiv preprint arXiv:2310.09909}, 2023.

\bibitem{46}
J.~J. Huallpa et~al.
\newblock Exploring the ethical considerations of using chat gpt in university education.
\newblock {\em Periodicals of Engineering and Natural Sciences}, 11(4):105--115, 2023.

\bibitem{47}
G.-G. Lee, E.~Latif, L.~Shi, and X.~Zhai.
\newblock Gemini pro defeated by gpt-4v: Evidence from education.
\newblock {\em arXiv preprint arXiv:2401.08660}, 2023.

\bibitem{48}
A.~M. Perlman.
\newblock The implications of chatgpt for legal services and society.
\newblock {\em Available at SSRN 4294197}, 2022.

\bibitem{21}
D.~M. Reddy, S.~M. Basha, M.~C. Hari, and N.~Penchalaiah.
\newblock Dall-e: Creating images from text.
\newblock {\em UGC Care Group I Journal}, 8(14):71--75, 2021.

\bibitem{50}
N.~Rane.
\newblock Role and challenges of chatgpt and similar generative artificial intelligence in arts and humanities.
\newblock {\em Available at SSRN 4603208}, 2023.

\bibitem{64}
S.~Yun, D.~Han, S.~J. Oh, S.~Chun, J.~Choe, and Y.~Yoo.
\newblock Cutmix: Regularization strategy to train strong classifiers with localizable features.
\newblock In {\em Proceedings of the IEEE/CVF International Conference on Computer Vision}, pages 6023--6032, 2019.

\bibitem{67}
Q.~Xie, Z.~Dai, E.~Hovy, T.~Luong, and Q.~Le.
\newblock Unsupervised data augmentation for consistency training.
\newblock {\em Advances in Neural Information Processing Systems}, 33:6256--6268, 2020.

\bibitem{82}
R.~Zhang, X.~Hu, B.~Li, S.~Huang, H.~Deng, Y.~Qiao, P.~Gao, and H.~Li.
\newblock Prompt, generate, then cache: Cascade of foundation models makes strong few-shot learners.
\newblock In {\em Proceedings of the IEEE/CVF Conference on Computer Vision and Pattern Recognition}, pages 15211--15222, 2023.

\bibitem{71}
G.~Chen, L.~Qiao, Y.~Shi, P.~Peng, J.~Li, T.~Huang, S.~Pu, and Y.~Tian.
\newblock Learning open set network with discriminative reciprocal points.
\newblock In {\em Computer Vision--ECCV 2020: 16th European Conference, Glasgow, UK, August 23--28, 2020, Proceedings, Part III}, pages 507--522. Springer, 2020.

\bibitem{72}
H.~Zhang, A.~Li, J.~Guo, and Y.~Guo.
\newblock Hybrid models for open set recognition.
\newblock In {\em Computer Vision--ECCV 2020: 16th European Conference, Glasgow, UK, August 23--28, 2020, Proceedings, Part III}, pages 102--117. Springer, 2020.

\bibitem{75}
J.~Lu, Y.~Xu, H.~Li, Z.~Cheng, and Y.~Niu.
\newblock Pmal: Open set recognition via robust prototype mining.
\newblock In {\em Proceedings of the AAAI Conference on Artificial Intelligence}, volume~36, pages 1872--1880, 2022.

\bibitem{76}
S.~Esmaeilpour, B.~Liu, E.~Robertson, and L.~Shu.
\newblock Zero-shot out-of-distribution detection based on the pre-trained model clip.
\newblock In {\em Proceedings of the AAAI Conference on Artificial Intelligence}, volume~36, pages 6568--6576, 2022.

\bibitem{77}
S.~Vaze, K.~Han, A.~Vedaldi, and A.~Zisserman.
\newblock Open-set recognition: A good closed-set classifier is all you need?
\newblock 2021.

\bibitem{79}
W.~Cho and J.~Choo.
\newblock Towards accurate open-set recognition via background-class regularization.
\newblock In {\em European Conference on Computer Vision}, pages 658--674. Springer, 2022.

\bibitem{81}
H.~Huang, Y.~Wang, Q.~Hu, and M.-M. Cheng.
\newblock Class-specific semantic reconstruction for open set recognition.
\newblock {\em IEEE Transactions on Pattern Analysis and Machine Intelligence}, 45(4):4214--4228, 2022.

\end{thebibliography}
\end{document}